\documentclass[lettersize,journal]{IEEEtran}
\usepackage{amsmath,amsfonts}
\usepackage{array}
\usepackage{textcomp}
\usepackage{stfloats}
\usepackage{url}
\usepackage{verbatim}
\usepackage{graphicx}
\usepackage{cite}

\usepackage{hyperref}
\usepackage{graphicx} 
\usepackage{amsmath} 
\usepackage{amssymb}  
\usepackage{tikz}
\usetikzlibrary{decorations.pathreplacing,calligraphy}
\usetikzlibrary{positioning}
\usepackage{caption}
\usepackage{subcaption}
\usepackage{makecell}
\usepackage{float}
\usepackage{rotating}
\usepackage{cite}
\usepackage{pgfplots}
\usepackage{tabularx, booktabs}
\usepackage{textgreek}
\usepackage{tcolorbox}
\usepackage[linesnumbered,ruled,vlined,Algorithm ]{algorithm }
\usepackage{algpseudocode}
\usepackage{booktabs}
\usepackage{multirow}
\usepackage{siunitx}
\usepackage{hyperref}
\usepackage{tabularx}
\usepackage{adjustbox}
\usepackage{wrapfig}

\begin{document}
\title{Predictive Maneuver Planning with Deep Reinforcement Learning (PMP-DRL) for comfortable and safe autonomous driving}

\author{Jayabrata Chowdhury, Vishruth Veerendranath, Suresh Sundaram, Narasimhan Sundararajan 
\thanks{Jayabrata Chowdhury is with Robert Bosch Centre for Cyber-Physical Systems, Indian Institute of Science, Bangalore, India. Suresh Sundaram  and Narasimhan Sundararajan are with the Artificial Intelligence and Robotics Lab, Dept. of Aerospace Engineering, Indian Institute of Science, Bangalore, India. Vishruth Veerendranath is with PES university, Bangalore, work done as a summer intern at Artificial Intelligence and Robotics Lab. {\tt\small Email: (jayabratac,vssuresh)@iisc.ac.in and (vishruthnath, ensundara)@gmail.com}}
\thanks{Manuscript received May 23, 2023}
}

\markboth{Journal of \LaTeX\ Class Files,~Vol.~14, No.~8, May~2023}%
{Shell \MakeLowercase{\textit{et al.}}: A Sample Article Using IEEEtran.cls for IEEE Journals}


\maketitle
\begin{abstract}
This paper presents a Predictive Maneuver Planning with Deep Reinforcement Learning (PMP-DRL) model for maneuver planning. Traditional rule-based maneuver planning approaches often have to improve their abilities to handle the variabilities of real-world driving scenarios. By learning from its experience, a Reinforcement Learning (RL)-based driving agent can adapt to changing driving conditions and improve its performance over time. Our proposed approach combines a predictive model and an RL agent to plan for comfortable and safe maneuvers. The predictive model is trained using historical driving data to predict the future positions of other surrounding vehicles. The surrounding vehicles' past and predicted future positions are embedded in context-aware grid maps. At the same time, the RL agent learns to make maneuvers based on this spatio-temporal context information. Performance evaluation of PMP-DRL has been carried out using simulated environments generated from publicly available NGSIM US101 and I80 datasets. The training sequence shows the continuous improvement in the driving experiences. It shows that proposed PMP-DRL can learn the trade-off between safety and comfortability. The decisions generated by the recent imitation learning-based model are compared with the proposed PMP-DRL for unseen scenarios. The results clearly show that PMP-DRL can handle complex real-world scenarios and make better comfortable and safe maneuver decisions than rule-based and imitative models.
\end{abstract}


\section{Introduction}
\IEEEPARstart{I}{ntelligent} Autonomous Vehicle (AV) development has advanced extraordinarily in recent years. However, autonomous driving on public roads still needs more work, especially in understanding the behaviors of surrounding traffic participants. To move safely and comfortably in dense traffic, an AV must understand the behaviors and intentions of its surrounding vehicles. One way to achieve this goal is a reliable vehicle communication system \cite{Bazzi2021AV_communication}. Using these communication channels, the AV can cooperate with other vehicles to decide its own actions. However, reliable communication can only be guaranteed sometimes. Another problem is that different vehicles have different communication protocols, and communicating is really challenging. To avoid this problem, another way is to make the AV utilize both past and present contextual information to understand the behaviors of surrounding vehicles within its sensor range. Even in this approach, some critical problems could be solved. One of them is handling uncertainties in human behavior. Another problem is that other human drivers' decisions influence the behavior of the primary ego vehicle's behavior. In multi-agent scenarios, the future context of one ego vehicle also depends on the other surrounding vehicles' future trajectories. In the context of an ego vehicle, the predicted future trajectories of surrounding vehicles can infer these intentions. From these predictions, a safe region can be formulated where the ego vehicle can move. Hence, there is a need for a planning strategy for an ego vehicle incorporating a prediction module for surrounding vehicles in predictive maneuver planning algorithms. This planning algorithm should also handle unseen scenarios, which is the main contribution of this paper.

Earlier works in this area have used techniques like Imitation Learning (IL), Reinforcement Learning (RL) for motion planning for an AV. Before presenting our approach, a brief review of the above methods is given below. In IL-based work \cite{bansal2018chauffeurnet}, a human-driven vehicle demonstration dataset has been used for learning the maneuver decision model. The learned driving model mimics a human driver's behavior. However, due to data distribution shifts from learning time, the IL model in \cite{bansal2018chauffeurnet} performed poorly in unknown scenarios like highly interactive lane change scenarios. In some work \cite{Pin2021AIRLdriving}, Inverse Reinforcement Learning (IRL) has been used to learn human driving behavior. In \cite{Pin2021AIRLdriving}, an IRL model Augmented Adversarial IRL (AugAIRL) was employed to change lanes in handcrafted highway scenarios. However, the absence of modeling uncertainties in the behaviors of surrounding vehicles in this paper represents a notable aspect with substantial implications for ensuring safety. Some models used open-source simulators like CARLA \cite{dosovitskiy2017carla} and Vista \cite{Amini_2022_Vista2}  for end-to-end motion planning for autonomous vehicles. In \cite{rhinehart2020deep_imitative_model}, an imitative model has been trained to understand desirable behavior for an interpretable expert-like driving using data collected from the CARLA simulator. In \cite{Codevilla2018ConditionalImit} and \cite{Cai2020Multimodal}, an imitation learning approach has been proposed for end-to-end autonomous driving. However, none of these models used trajectory predictions for evaluating behavior uncertainties. Recent works \cite{shu2022short_term_traffic}, \cite{katariya2022Deep_track}, \cite{Mersch2021spatio} use deep neural networks for trajectory predictions. These works model complex spatio-temporal relationships among the different vehicles. DST-CAN \cite{Jayabrata2022DST_CAN} has recently modeled the prediction uncertainties in a maneuver planning model. DST-CAN used an imitation model for interactive highway driving and showed the effectiveness of the use of a trajectory prediction module for safe maneuver planning for real-world scenarios using NGSIM I80 and US101 \cite{colyar2007us101} datasets. However, since DST-CAN is an imitative model, it only tried to mimic some human behavior and suffered from scenarios where passenger comfort is also required along with safety.

An RL agent that learns through its own experience can overcome an imitative model's data distribution shift limitations. The works in 	\cite{chen2019model}, \cite{chen2021interpretable} used RL for learning neural network models to make maneuver decisions. These works use Bird's Eye View (BEV) raster image-based models for all the surrounding vehicles. This BEV representation has served as the input to the policy network. In some other works (\cite{ye2020automated}, \cite{tang2022highway}), policy optimization methods were applied for lane-changing maneuvers. The work in \cite{ye2020automated} employs a Proximal Policy Optimization (PPO) \cite{schulman2017PPO} algorithm in a simulated environment for learning discrete lateral and longitudinal actions. In \cite{tang2022highway}, a Soft Actor-Critic (SAC) \cite{Harnoja2018SAC} algorithm learns how to make steering and acceleration decisions in handcrafted traffic scenarios. In \cite{Alizadeh2019LaneChangeHighway}, \cite{Kaleb2021HDQN}, Deep Q Networks (DQNs) were used for lateral and longitudinal decisions. However, these works must incorporate the prediction module for surrounding vehicles to understand the future context and safe and comfortable decision-making. A detailed survey of RL-based autonomous driving can be found in \cite{survey_RL_for_AD}. Another concern is that RL learns in an active learning framework. Each time the RL agent interacts with the environment, it must collect new experiences to update the model. Getting real-world data during RL exploration may only be safe sometimes, as it can take dangerous actions. The surrounding traffic participants follow some hand-engineered rules in a closed simulation environment. The recent advancements in other fields of deep learning, like computer vision and natural language processing, have come from the help of vast and diverse real-world datasets. The RL algorithms can perform much better if some real-world datasets with interactions among vehicles are used for learning. Recent works \cite{levine2020offline}, \cite{prudencio2022survey} focus on the importance of data-driven Offline Reinforcement Learning (ORL) frameworks. Based on the real-world data-based Vista simulator, \cite{Amini_2020_robust_control} applied deep reinforcement learning to learn a policy to drive on unseen scenarios. However, this work omitted the incorporation of free space prediction and exclusively relied on the sparse reward to assess performance. An RL-based driving model with sparse reward could not learn to drive in a dense interactive scenario like a highway because there should be some signal to show how the RL agent performs every step. Hence, there is a need to develop a predictive maneuver planning algorithm with RL that can learn from real-world interaction and make both safe and comfortable decisions.

This paper proposes a new Predictive Maneuver Planning with Deep Reinforcement Learning (PMP-DRL) approach in AVs for safe and comfortable maneuver actions. The proposed approach assumes that a state-of-the-art perception module tracks other vehicles inside AV's sensor range and gives the positions of the vehicles. After observing the past positions, or the past trajectories, of the surrounding vehicles, a recurrent Memory Neuron Network (MNN) \cite{sastry1994memory, helicopter_dynamics} predicts the future trajectories of those vehicles without any HD map, learning only with vehicle dynamics. These predicted trajectories of the surrounding vehicles are passed through a context generator. The context generator generates the input representation for spatio-temporal context-aware grid maps. The grid maps contain the past, present, and predicted future contextual information. The probabilistic occupancy grid maps encode future positions with prediction uncertainties. This work uses real-world driving interaction datasets for RL agents. NGSIM I80 and US-101 \cite{colyar2007us101} datasets are used to learn and evaluate the learned RL agent. A simulation environment has been developed where surrounding vehicles move according to the trajectories given in the datasets. Some reward function designs for a dense reward function and network architecture tricks have been developed based on the dataset for better performance. The Double Deep Q Network (DDQN) \cite{van2016DDQN} is employed to learn the hierarchical lateral and longitudinal decisions with the unicycle model for continuous decisions. The model was trained with high and low-congestion datasets and evaluated with the medium-congestion dataset as given in DST-CAN \cite{Jayabrata2022DST_CAN}. This paper gives importance to passenger comfort along with safety. Hence, the experiments are developed to find passenger comfort in the number of cases with jerkings and the safety of actions taken. The training sequence consistently improves average acceleration, passenger comfortability, and safety measures. The final comparative results show that our proposed PMP-DRL performs  $51.79\%$  and $51.20\%$ better than the rule-based and baseline imitation-based model in unseen traffic scenarios. It has been found that there is a trade-off between safety and passenger comfort.  

The paper is organized as given below. Section \ref{PMP-DRL} describes the methodology for solving the problem of predictive motion planning with RL, details of action space, reward design, and architectural details. Section \ref{Performance evaluation of PMP-DRL} details the simulation environment, metrics developed, and performance evaluation of PMP-DRL. Finally, the conclusions from the study are summarised in section \ref{conclusions}.

\section{Predictive Maneuver Planning with Deep Reinforcement Learning (PMP-DRL)}
\label{PMP-DRL}
This section describes the method to solve Predictive Maneuver Planning with Deep Reinforcement Learning (PMP-DRL) scheme. The approach uses Double Deep Q Network (DDQN) with context-aware grid maps processed by a Convolutional Neural Network (CNN). Before getting into the details of the architecture and training of the algorithm for PMP-DRL, the design of a context-aware gird-based observation space, action space design for comfort and safety analysis, and design of dense reward function are discussed.
\begin{figure}[h!]
\centering
\includegraphics[width=3.4in]{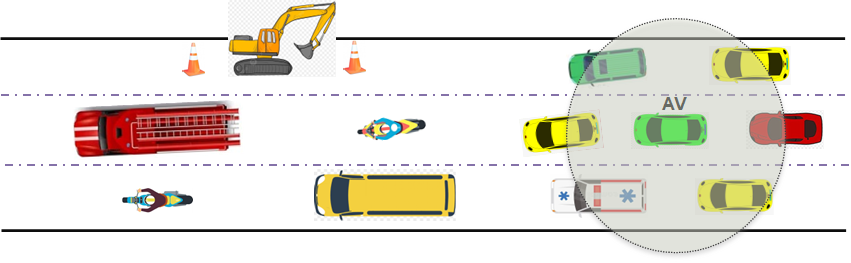}
\caption{A common traffic scenario with an AV in operation}
\label{figure1}
\end{figure}

\subsection{Problem  formulation}
\begin{figure}[h!]
\centering
\includegraphics[width=3.4in]{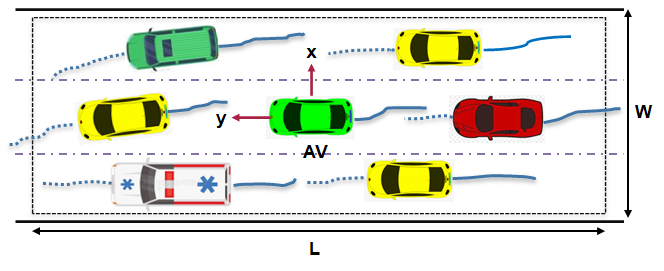}
\caption{A context region with an AV at the center}
\label{context_region}
\end{figure}
One of the biggest challenges in safe and comfortable driving on a public road is understanding the behavior of the ego AV's surrounding vehicles. If the surrounding vehicles' future poses can be predicted, then it is easier to understand their behavior. Fig. \ref{figure1} shows the importance of predicting the future positions of the surrounding vehicles. The dotted circular line shows the AV's sensor range. Since the static obstacle is not in the sensor range of the AV,  it cannot detect the reason for the traffic flow change. If surrounding vehicles' future positions can be predicted, it will be easier for the AV to understand how the driving scenario will evolve. AV can then decide on a safe and passenger-comfortable action depending on the predicted trajectories of surrounding vehicles. In recent work \cite{Alizadeh2019LaneChangeHighway}, Deep Reinforcement Learning (DRL) has been used for decision-making. This paper employs DRL for maneuver decision-making, incorporating a prediction module for safe and comfortable decisions.  

\subsection{Observation space for DRL}
This paper presents a way to incorporate the spatio-temporal context information of an AV in an occupancy grid-based format without using High Definition map. We assume that HD map information is unavailable to an AV since building and maintaining a HD map is difficult. Also, HD map information processing is computationally heavy, and it may be noted that all sensors have their limitations. Hence, in this paper, context-aware grid maps are developed around the AV. These grid maps consist of occupancy maps for the present and past time frames and Probabilistic Occupancy Maps (POMs) with grid occupancy probabilities of surrounding vehicles. Depending on the AV's sensor range, a context region is chosen for making the context-aware grids, as shown in Fig.\ref{context_region}. The context region assumes two adjacent lanes and a sensor range of 90 feet for the AV, as shown in Fig.\ref{context_region}. ${\textbf{P}_o}$ and ${\textbf{P}_e}$ denote the past positions of the surrounding vehicles and the ego vehicle for $p$ time frames. For Predictive Motion Planning (PMP), future positions of the surrounding vehicles' are predicted as ${\hat{\textbf{P}}_o}$ for $T$ future timesteps. 
\begin{equation}
\begin{aligned}
    \{\textbf{P}_e\}=\{(x^{-p}_a,y^{-p}_a),(x^{-p+1}_a,y^{-p+1}_a),...,(x^0_a,y^0_a)\} \\
    \{\textbf{P}_o\}=\{(x^{-p}_{o_{i}},y^{-p}_{o_{i}}),(x^{-p+1}_{o_{i}},y^{-p+1}_{o_{i}}),...,(x^0_{o_{i}},y^0_{o_{i}})\} \\
    \{\hat{\textbf{P}}_o\}=\{(\hat{x}^{1}_{o_{i}},\hat{y}^{1}_{o_{i}}),(\hat{x}^{2}_{o_{i}},\hat{y}^{2}_{o_{i}}),...,(\hat{x}^T_{o_{i}},\hat{y}^T_{o_{i}})\}
\end{aligned}
\end{equation}
where $i=1,2,...,N$; \textit{N} is number of surrounding vehicles inside AV's sensor range. 
\begin{figure}
    \centering
    \begin{subfigure}[b]{0.15\textwidth}
        \centering
        \includegraphics[width=\textwidth]{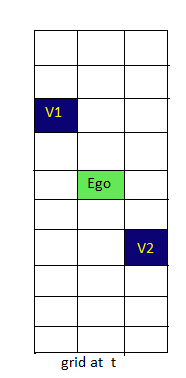}
        \caption{context grid map}
        \label{grid_t}
    \end{subfigure}
    \hfill
    \begin{subfigure}[b]{0.15\textwidth}
        \centering
        \includegraphics[width=\textwidth]{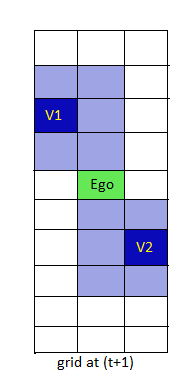}
        \caption{POM}
        \label{grid_t+1}
    \end{subfigure}
    \hfill
    \begin{subfigure}[b]{0.15\textwidth}
        \centering
        \includegraphics[width=\textwidth]{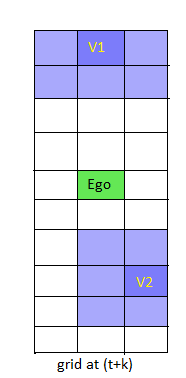}
        \caption{POM}
        \label{grid_t+i}
    \end{subfigure}
    \caption{Context and POM embedding for different time steps}
    \label{Context-POM-embedding}
\end{figure}
Here, we use a Memory Neuron Network (MNN) for the trajectory prediction of vehicles as described in \cite{Rao2021Spatio}. We assume that HD map information is unavailable to an AV since building and maintaining an HD map is difficult. Also, HD map information processing is computationally heavy, and all sensors have limitations. Hence, the MNN is trained with Root Mean Squared Error (RMSE) loss function for trajectory prediction using only position information. MNN has been employed for multi-step prediction of future positions as given in the Eqns \ref{eq:increment1} and \ref{eq:increment2}.
\begin{equation}
\begin{aligned}
    \bar{x}_{t+1} = x_t + \Delta \bar{x}_t ; \ 
    \bar{y}_{t+1} = y_t + \Delta \bar{y}_t
\end{aligned}
\label{eq:increment1}
\end{equation}
\begin{equation}
    \begin{aligned}
    \bar{x}_{t+(k+1)} = \bar{x}_{t+k} + \Delta \bar{x}_{t+k} ; \ 
    \bar{y}_{t+(k+1)} = \bar{y}_{t+k} + \Delta \bar{y}_{t+k}
\end{aligned}
\label{eq:increment2}
\end{equation}
where $k = 1,2,...,(T-1)$. Occupancy grid maps are then created from the predicted positions of surrounding vehicles with respect to the distance from the ego vehicle. Since there are always uncertainties in predicting surrounding vehicles' future positions with increasing prediction horizons, it has been assumed here that a vehicle can be at any of the surrounding eight grids around the grid predicted by MNN. This assumption comes from the RMSE for different prediction horizons. The predicted grid by MNN is given an occupancy probability value as given in Equ. \ref{eq:decay}. 
\begin{equation}
    \textbf{P}(t) = 0.47 + \sqrt{0.236 - 0.004t}
    \label{eq:decay}
\end{equation} where $0\le t \le 30$ is the time index.
\begin{algorithm}
\begin{algorithmic}[1]
\State {\textbf{Inputs}: Past track history set $\{\mathbf{X}_h\}_{i=1}^{\mathcal{D}}$ where $\mathcal{D}$ is the number of surrounding vehicles detected by the AV sensors, pre-trained weights for Memory Neuron Network (MNN), current time step $t$}, prediction horizon $T$
\State {\textbf{Initialize}: $\mathcal{G}$ is a $13\times3\times60$ array of zeros}
\State {\textbf{Step 1}: Obtain the \textit{context-aware grid $\mathcal{G}$}:}
\For{$v \leftarrow 1$ to $\mathcal{D}$}
\For{$\tau \leftarrow t-30$ to $t+T$}
\If{$\tau \leq$ t}
\State{$\mathcal{G}(i, j, \tau) = 1.0$, where $(i, j)$ is the grid \newline \phantom a \phantom a \phantom a\phantom a indices obtained using the relative position $\mathbf{X}_h$ \newline \phantom a \phantom a \phantom a   \phantom. of vehicle $v$ w.r.t ego vehicle's position at time $\tau$.}
\Else 
\State{$\mathcal{G}(i, j, \tau) = \mathbf{P}(\tau)$, where $(i, j)$ is the grid  \newline \phantom a \phantom a \phantom a \phantom a indices obtained using the predicted position of \newline \phantom a \phantom a \phantom a \phantom . vehicle $v$ (from MNN), and $\mathbf{P}(\tau)$ is calculated \newline \phantom a \phantom a \phantom a \phantom . using \ref{eq:decay}}. $\mathcal{G}(i\pm 1, j\pm 1, \tau) = \mathcal{G}(i\pm 1, j, \tau) = \mathcal{G}(i, j\pm 1, \tau) = ((1 - \mathbf{P}(\tau))/8$
\EndIf
\EndFor
\EndFor
\caption{Context-aware grids generation pseudo code}
\label{Context generation code}
\end{algorithmic}
\end{algorithm}
This study used $3$ seconds of past trajectories to build context-aware grid maps as DST-CAN \cite{Jayabrata2022DST_CAN} showed that $3$ seconds prediction horizon is good enough for planning. One occupancy grid map and POM are shown in Fig. \ref{Context-POM-embedding}. The final past occupancy maps and POMs are stacked to make a $3$-dimensional context-aware occupancy grid map consisting of 60 channels of 13x3 dimensions. The occupancy grid map creation uses the same procedure as described in DST-CAN \cite{Jayabrata2022DST_CAN} and is given in Algo \ref{Context generation code}.

\subsection{Action space}
The PMP-DRL algorithm aims to generate safe, efficient, and comfortable maneuvering decisions from predicted trajectories of surrounding vehicles. The action space design takes care of the ego vehicle passenger comfort. A sudden change in maneuver decisions is not desirable for passengers. Hence, action space should be defined in such a way that only whenever required, the ego vehicle brakes hardly or suddenly changes lanes. Hence, the following abstract-level discrete meta-actions \textit{lateral maneuver = [hard left, soft left, same lane, soft right, hard right]} and \textit{longitudinal maneuver = [accelerate, cruise, decelerate, brake]} are chosen to consider as the action space. This action space should reflect the comforts of passengers and the decisions like \textit{braking} should be taken only when there is no way to prevent near-collision scenarios. If enough space is available in front of the ego vehicle, it should only slowly decelerate because it is comfortable for passengers. The discrete meta-actions are then converted to continuous action signals for taking a step in the driving environment. There is a fixed change in velocity and yaw angle for each action. Since accurately predicting the bicycle model parameters are difficult with vision sensors' limitations, the ego vehicle uses a unicycle model as given in Equ. \ref{unicycle_model}. 
\begin{equation}
\begin{aligned}
    \dot{x}_t = v_t * \sin \phi_t ; \ \dot{y}_t = v_t * \cos \phi_t ; \ \dot{\phi}_t = \Delta{\phi}_t
\end{aligned}
\label{unicycle_model}
\end{equation}
where $x_t, y_t, \phi_t$ are the positions and yaw angle of the ego vehicle. $v_t$ is the velocity of the ego vehicle at the current time instant. The discrete lateral actions have been converted to continuous changes in the ego vehicle's yaw angles \textit{$\Delta \phi_t$}. Similarly, longitudinal actions are converted to changes in the ego vehicle's velocity \textit{$\Delta v_t$}.
\begin{equation}
\begin{aligned}
    v_t^x  = (x_t - x_{t-1})/\Delta t ; \ v_t^y  = (y_t - y_{t-1})/\Delta t \\
    v_t = \sqrt{(v_t^x)^2 + (v_t^y)^2};\ \phi_t = \tan^{-1} (x_t - x_{t-1}/y_t - y_{t-1}) \\
    v_{up} = v_t + \Delta v_t; \ \phi_{up} = \phi_t + \Delta \phi_t \\ x_{t+1} = x_t + (v_{up}\sin \phi_{up}\Delta t);\ y_{t+1} = y_t + (v_{up}\cos \phi_{up}\Delta t) 
\end{aligned}
\label{update_equ}
\end{equation}
Since the action space is discrete, the discrete actions are converted to a change in control action \textit{$u_t=[\Delta v_t, \ \Delta \phi_t]$} for updating the position of ego vehicle with the unicycle model. In Equ. \ref{update_equ}, $x_t$, $y_t$ are positions of the ego vehicle at the current time instant. $x_{t-1}$, $y_{t-1}$ are position of ego vehicle at previous time instant. $\Delta t$ is the time interval. In this work, $\Delta t = 0.1$ seconds. In Equ. \ref{update_equ}, at first current instant velocity $v_t$ and yaw angle $\phi_t$ are calculated. Next, velocity and yaw angle is updated to get new velocity $v_{up}$ and yaw angle $\phi_{up}$. Equ  \ref{update_equ} then calculates the next position of the ego vehicle.

\subsection{Reward function}
\begin{figure}
\centering
\begin{adjustbox}{max size={.3\textwidth}{.3\textheight}}
\begin{tikzpicture}
\draw (0,0) -- (8,0) -- (8,8) -- (0,8) -- (0,0);
\draw[violet, thick, dashed] (1,0) -- (1,8);
\draw[violet, thick, dashed] (3,0) -- (3,8);
\draw[violet, thick, dashed] (5,0) -- (5,8);
\draw[violet, thick, dashed] (7,0) -- (7,8);
\draw[black, dashed] (4,4) circle(3.7cm);
\draw[black, dashed] (4,4) circle(2.5cm);
\draw (0,2.85) -- (8,2.85);
\draw (0, 5.15) -- (8,5.15);
\draw[blue, thick] (3.5702, 2.95) -- (4.4298, 2.95) -- (4.4298, 5.0325) -- (3.5702, 5.0325) -- (3.5702, 2.95);
\fill[green](3.5702, 2.95)--(4.4298, 2.95)--(4.4298, 5.0325)--(3.5702, 5.0325);
\fill [blue] (4,4) circle (2pt);
\draw [blue] (4,4) -- (4,7.7);
\node [blue, rotate= -90] at (4.3, 7.0) {$d_2$};
\draw [blue] (4, 4) -- (6.22, 5.15);
\node [blue, rotate= 25] at (5.7, 4.7) {$d_1$};
\node at (4.0, -0.3) {Ego lane};
\node at (2.0, -0.3) {Left lane};
\node at (6.0, -0.3) {Right lane};
\node [red] at (3.5, 6.0) {N};
\node [red] at (3.3, 4.3) {N};
\node [red] at (4.7, 4.0) {N};
\node [red] at (4.5, 2.0) {N};
\node [black!66!green] at (6.0, 4.0) {$P_1$};
\node [black!66!green] at (2.5, 4.5) {$P_1$};
\node [black!66!green] at (1.5, 6.0) {$P_2$};
\node [black!66!green] at (6.5, 6.0) {$P_2$};
\node [black!66!green] at (5.5, 5.4) {$P_2$};
\node [black!66!green] at (2.5, 5.4) {$P_2$};
\node [black!66!green] at (2.5, 2.6) {$P_2$};
\node [black!66!green] at (5.5, 2.6) {$P_2$};
\node [black!66!green] at (1.5, 2.0) {$P_2$};
\node [black!66!green] at (6.5, 2.0) {$P_2$};
\node [black!66!green] at (3.5, 1.0) {$P_2$};
\node [black!66!green] at (1.5, 0.5) {$P_3$};
\node [black!66!green] at (6.5, 0.5) {$P_3$};
\node [black!66!green] at (1.5, 7.5) {$P_3$};
\node [black!66!green] at (6.5, 7.5) {$P_3$};
\node [blue, rotate=90] at (-0.4, 4.5) {$0.5l$};
\draw [black, dashed] (0, 4.0) -- (4, 4.0);
\draw [blue] (4.0, 1.5) -- (5.0, 1.5);
\draw [black, thick, dotted] (4,4) -- (4.0, 1.5);
\node [blue] at (4.5, 1.3) {$0.5l$};
\end{tikzpicture}
\end{adjustbox}
\caption{Positive and negative reward region for AV}
\label{Reward region}
\end{figure}
\begin{algorithm}
\begin{algorithmic}[1]
\State {\textbf{Input}: Predicted future position ($ego_x^{pr}$, $ego_y^{pr}$) of ego vehicle for action, actual ego vehicle position ($ego_x^{ac}$, $ego_y^{ac}$) in future extracted from dataset, surrounding vehicles' future positions ($s_x^i$, $s_y^i$) where $i=1,2,...,S$; $S$ is number of surrounding vehicles in sensor range.}
\State {\textbf{Output}: Total reward for an action}
\State {\textbf{Initialize}: $p_1=0, p_2=0, p_3=0, p_{count}=0, n_{count}=0, r_{dis}=0, r_{pos}=0, r_{neg}=0$, $r_{imit}=0$, $r_{off-road}=0$, $reward=0$, $c_1=5, c_2=125, k_1=2, k_2=-6$}
\State {\textbf{Part 1:} Distance based reward}
\For{$i \leftarrow 1$ to S}
\State $\Delta x = ego_x^{pr} - s_x^i; \Delta y = ego_y^{pr} - s_y^i; d = \sqrt{(\Delta x) ^{2} + (\Delta y)^{2}}$
\If{$|\Delta y| \leq 0.5l$}
\If{$|\Delta x| \geq 0.5l$}
\State $p_1 \leftarrow p_1 + 1$; $p_{count} \leftarrow p_{count}+1$
\If{$p_1 \leq 1$}
\State $r_{pos} = r_{pos} + c_1 \tanh{(|\Delta x| - 0.5l)}$
\EndIf
\Else
\State $n_{count} = n_{count} + 1$
\State $r_{neg} = r_{neg} + c_1 \tanh{(|\Delta x| - 0.5l)}$
\EndIf
\Else
\If{$d \leq d_1$ and $|\Delta| x \leq 0.5l$}
\State $n_{count} = n_{count} + 1$
\State $r_{neg} = r_{neg} + c_1 \tanh{(d - d_1)}$
\ElsIf {$d \leq d_2$}
\State $p_2 \leftarrow p_2 + 1$; $p_{count} \leftarrow p_{count}+1$
\State $r_{pos} = r_{pos} + (d/c_1)$
\Else
\State $p_3 \leftarrow p_3 + 1; p_{count} \leftarrow p_{count}+1$
\State $r_{pos} = r_{pos} + (c_2/d)$
\EndIf
\EndIf
\EndFor
\State $r_{dis} = r_{dis} + (r_{pos}/p_{count})+ k_1 r_{neg}$
\State \textbf{Part 2:} Imitation based reward
\State $x_{err} = ego_x^{pr} - ego_x^{ac}; \ y_{err} = ego_y^{pr} - ego_y^{ac}$
\State $imit_{err} = 0.25|x_{err}| + 0.1|y_{err}|; \ r_{imit} = -0.5 (imit_{err})$
\State \textbf{Part 3:} Off-road reward
\State \textbf{if} $ego_x^{pr} \leq 0 \ or  \ ego_x^{pr} \geq (\#lane * lane \ width) \textbf{then} r_{off-road} = k_2$
\State $reward = r_{dis} + r_{imit} + r_{off-road}$ 
\caption{Details of Reward function}
\label{Reward function}
\end{algorithmic}
\end{algorithm}
The reward function is a crucial component of DRL algorithms. Generally, a positive reward is given if an AV reaches its goal. However, training a DDQN with this kind of sparse reward is difficult. Hence, the reward has been modified using distance-based, imitation-based, and off-road components. One component is a distance-based reward for preventing near-collision scenarios. The second part is a reward for imitation learning based on human driving behavior. The third and last part is an off-road reward for preventing the ego vehicle from taking an action that can go out of the road boundary. These three components are then combined to contribute to the final reward for an action. The safe region around the ego vehicle is defined based on the average length ($l=15 foot \approx 4.57 m$) and width ($w$) of vehicles on the road given in the dataset. If any vehicle comes near the ego vehicle laterally, a negative reward is given for that action. If any surrounding vehicle is within distance $d_1$ of the ego vehicle on the front and backside, there is a risk of collision. Hence, this region is a negative reward region. The ego vehicle is at the center, shown in green in Fig.\ref{Reward region}. The positive and negative reward regions have been shown in Fig.\ref{Reward region} with "P" and "N." The positive reward region is divided into three parts, as shown in Fig. \ref{Reward region}. If a vehicle is moving in the same direction parallelly to the ego vehicle, maintaining a safe distance (within distance $d_1=(l+1) foot = 16 foot \approx 4.88 m$), then that region is denoted as positive region $P_1$. If a vehicle is not very close to the ego vehicle (within distance $d_2=(1.5l+2.5) foot = 25 foot \approx 7.62 m$), then it is in the positive reward region $P_2$. If any vehicle is more than $d_2$ distance from the ego vehicle, then that region is a positive reward region $P_3$. The number of surrounding vehicles in each region is calculated as $p_1, p_2, p_3, n_{count}$ to define the final distance-based reward. The reward is designed for a smooth transition from one region to another. The reward is then scaled so that vehicle in a near-collision scenario gets higher importance. The imitation part of the reward is calculated by the distance of the ego vehicle from the position of human decisions. The imitation part of the reward is scaled according to the $x,y$ position displacement error. The reward design details are given in Algo.\ref{Reward function}.

\subsection{Architecture and training details}
\begin{figure*}[h!]
    \begin{tikzpicture}
        \draw (0, -4.8) rectangle (2.0, 0.2);
        \node at (1.0, 0.0) {Simulation};
        \node at (1.0, -0.4) {environment};
        \node[inner sep=0pt] (current) at (1.1, -2.6){\includegraphics[width=1.5 cm, height=3.8cm]{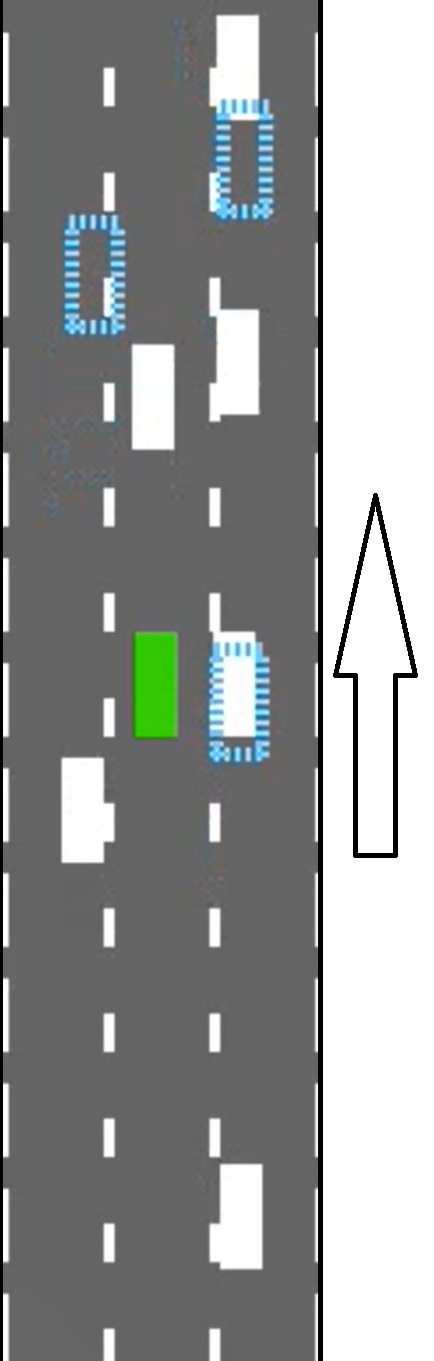}};
        
        \draw[-latex, thick] (1.0, 0.2) -- (1.0, 3.0);
        \draw (0,3) rectangle (2.0, 4.5);
        \node at (1.0, 4.0) {Obs,};
        \node at (1.0, 3.6) {reward};
        \node at (7.0, 4.0) {Past trajectories of surrounding traffic participants};
        \draw[-latex, thick] (2.0, 3.75) -- (13.0, 3.75);
        
        \draw (13.0, 2.0) rectangle (18.0, 4.5);
        \node[inner sep=0pt] (current) at (16.5, 3.3){\includegraphics[width=2.0 cm, height=2.0cm]{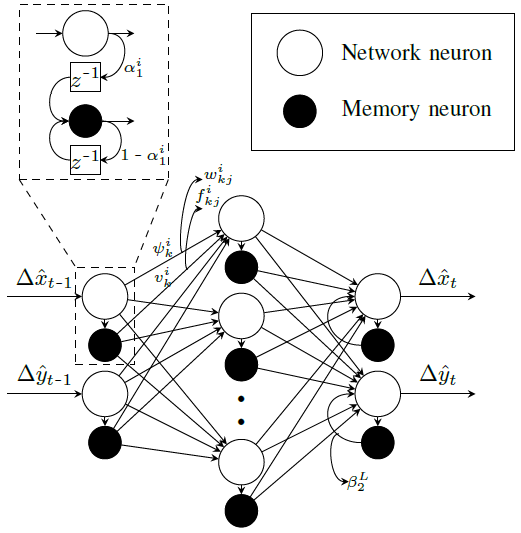}};
        \node at (14.0, 4.0) {Context};
        \node at (14.0, 3.4) {generator};
        \node at (14.0, 2.8) {with MNN};
        \node[inner sep=0pt] (current) at (10.5, 2.8){\includegraphics[width=4.5 cm, height=1.8cm]{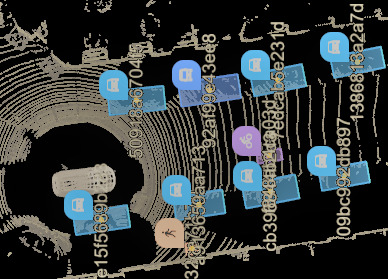}};
        \node[inner sep=0pt] (current) at (5.5, 2.8){\includegraphics[width=5.0 cm, height=1.0cm]{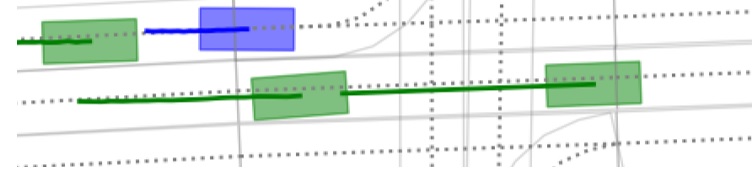}};
        \draw[-latex, thick] (16.0, 2.0) -- (16.0, 0.8);
        
        \draw (14.0, -3.5) rectangle (18.0, 0.8);
        \node at (16.0, 0.57) {Context-aware grid map};
        \node[inner sep=0pt] (current) at (15.0, -1.5){\includegraphics[width=1.5 cm, height=3.8cm]{Images/grid_t.png}};
        \node[inner sep=0pt] (current) at (17.0, -1.5){\includegraphics[width=1.5 cm, height=3.8cm]{Images/grid_t+k.png}};
        \draw[-latex, thick] (14.0, -1.25) -- (12.6, -1.25);
        
        \draw (4.0, -6.5) rectangle (13.0, 0.8);
        \node at (8.0, 1.0) {Action engine};
        \draw (11.6, -2.0) rectangle (12.6, -0.5);
        \node at (12.1, -1.0) {\small$3\times3$};
        \node at (12.1, -1.5) {Conv};
        \draw[-latex, thick] (11.6, -1.25) -- (11.1, -1.25);
        \draw (10.1, -2.0) rectangle (11.1, -0.5);
        \node at (10.6, -1.0) {\small$3\times1$};
        \node at (10.6, -1.5) {Conv};
        \draw[-latex, thick] (10.1, -1.25) -- (9.5, -1.25);
        \draw (8.6, -2.0) rectangle (9.5, -0.5);
        \node at (9.1, -1.0) {Max};
        \node at (9.1, -1.5) {Pool};
        \draw[-latex, thick] (8.6, -1.25) -- (8.0, -1.25);
        \node[inner sep=0pt] (current) at (7.2, -1.1){\includegraphics[width=1.5 cm, height=3.6cm]{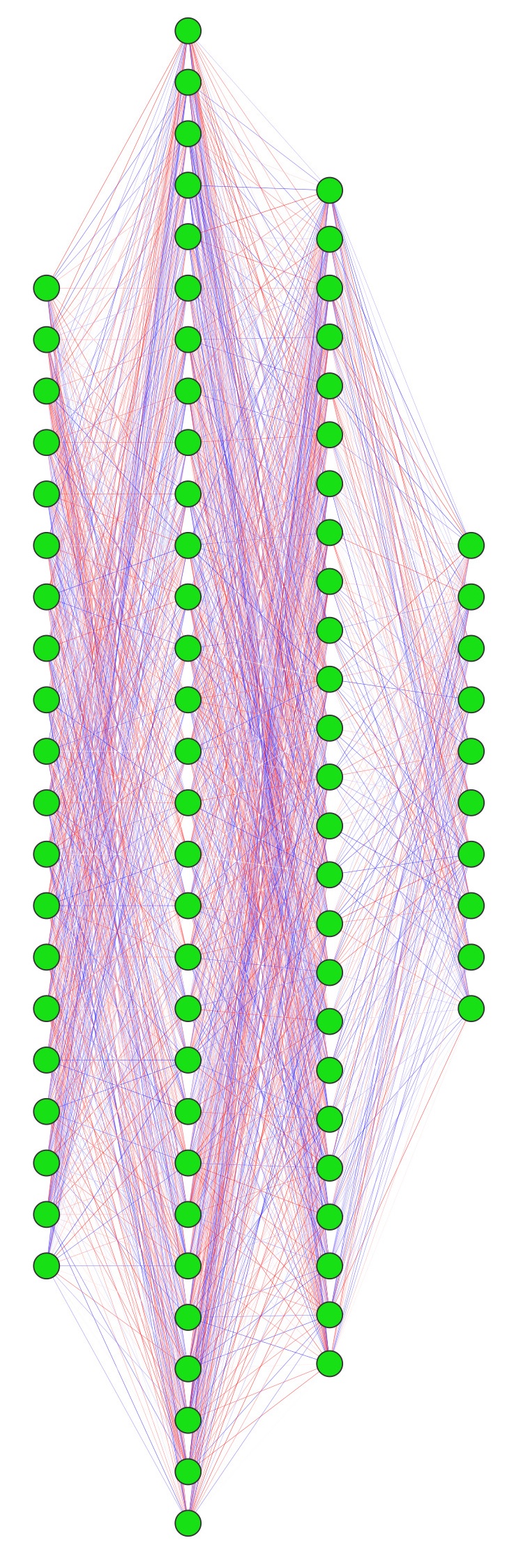}};
        \draw [blue, decorate, decoration = {brace}] (6.5, -2.3) -- (6.5, 0.1);
        \draw[blue, -latex, thick] (6.4, -1.1) -- (5.8, -1.1);
        \draw (4.5, -1.5) rectangle (5.8, -0.3);
        \node at (5.2, -0.9) {Argmax};
        \node at (5.0, 0.3) {Q Network};
        \draw[blue, -latex, thick] (4.5, -1.0) -- (2.0, -1.0);
        \node at (3.2, -0.5) {DDQN};
        \node at (3.2, -0.8) {Action};
        \draw [pen colour={orange}, decorate, decoration = {calligraphic brace, mirror, raise=5pt, amplitude=5pt}] (8.5, -2.0) -- (12.7, -2.0);
        \node at (10.6, -2.5) {State encoding};
        \node[inner sep=0pt] (current) at (11.2, -4.5){\includegraphics[width=3.3 cm, height=1.5cm]{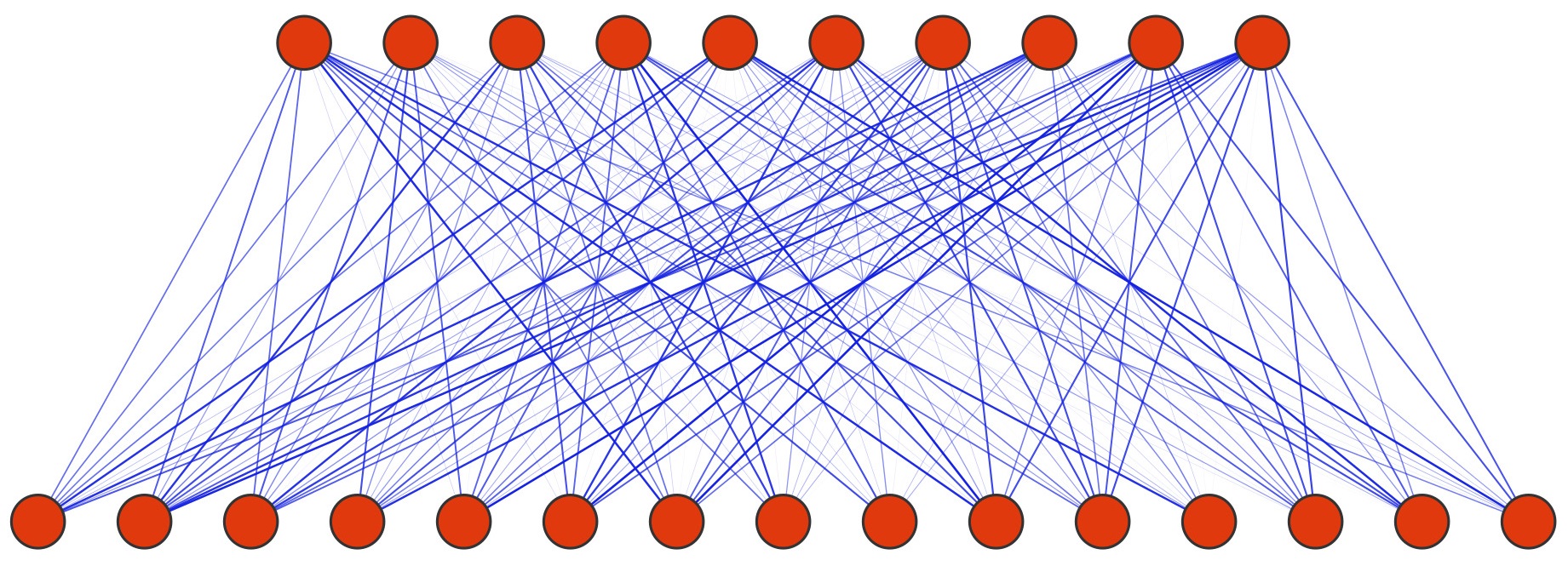}};
        \node[inner sep=0pt] (current) at (5.2, -3.2){\includegraphics[width=1.0 cm, height=2.0cm]{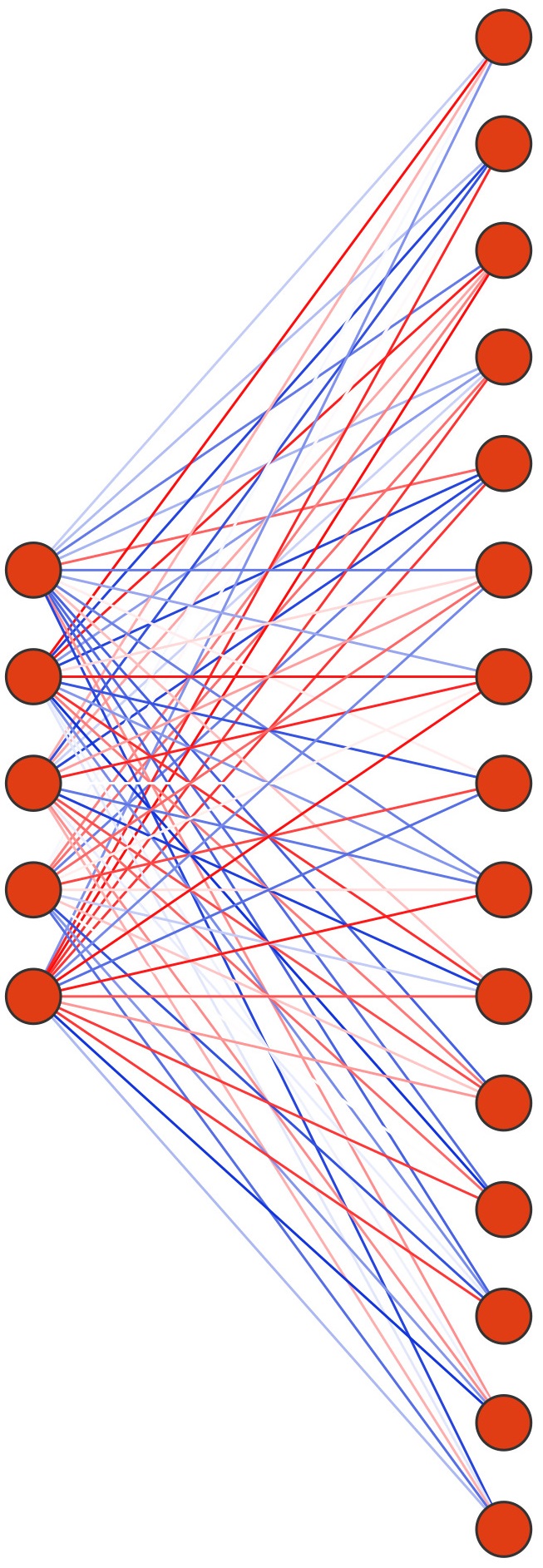}};
        \node[inner sep=0pt] (current) at (5.2, -5.4){\includegraphics[width=1.0 cm, height=2.0cm]{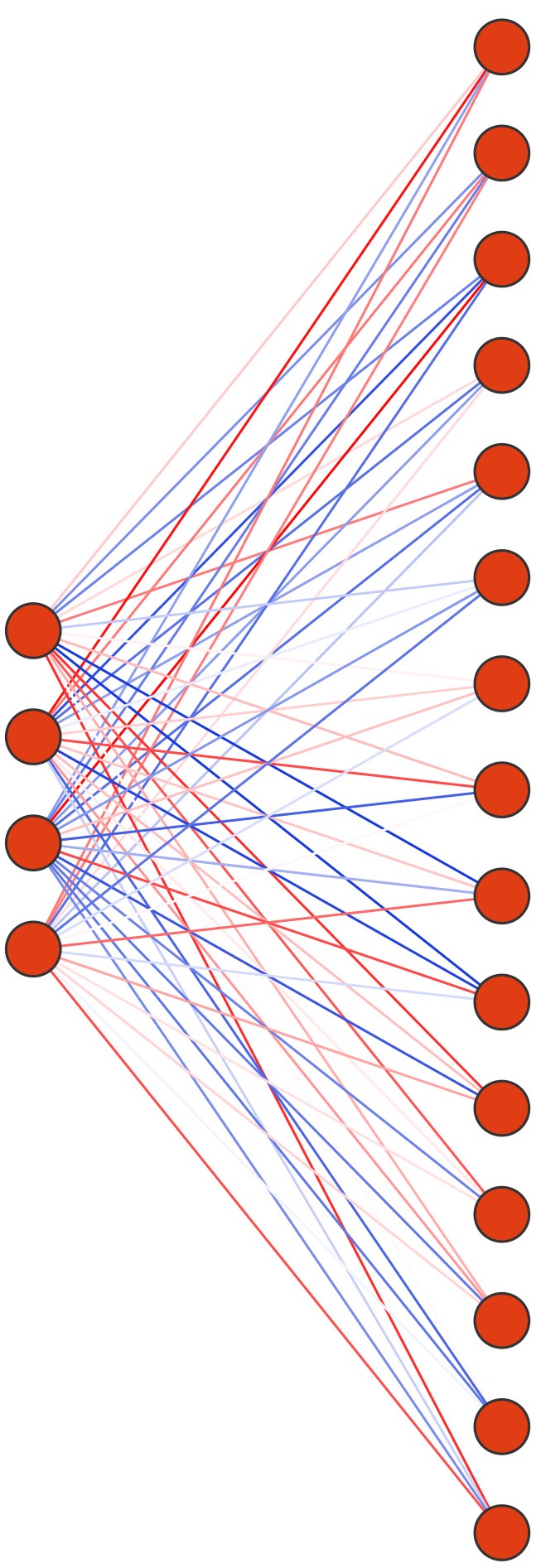}};
        \draw[thick] (8.3, -1.25) -- (8.3, -3.25);
        \draw[thick] (8.3, -3.25) -- (11.0, -3.25);
        \draw[-latex, thick] (11.0, -3.25) -- (11.0, -3.8);
        \draw[red, thick] (11.0, -5.3) -- (11.0, -5.8);
        \draw[red, -latex, thick] (11.0, -5.8) -- (5.8, -5.8);
        \draw[red, thick] (7.0, -5.8) -- (7.0, -3.5);
        \draw[red, -latex, thick] (7.0, -3.5) -- (5.8, -3.5);
        \node at (8.5, -4.2) {Imitation network};
        \draw[red, -latex, thick] (4.5, -3.2) -- (3.0, -3.2);
        \draw[red, thick] (4.5, -5.4) -- (3.0, -5.4);
        \draw[red, -latex, thick] (3.0, -5.4) -- (3.0, -3.2);
        \draw[red, -latex, thick] (3.0, -3.2) -- (2.0, -3.2);
        \node at (3.2, -2.7) {Imitation};
        \node at (3.2, -3.0) {Action};
    \end{tikzpicture}
\caption{Architecture for PMP-DRL}
\label{schematic_dia}
\end{figure*}
The architecture of the proposed PMP-DRL method is shown in Fig.\ref{schematic_dia}. A simulation environment has been developed based on a real-world NGSIM dataset. At every time step, the simulation environment will reward the ego vehicle and provide observation containing information about the positions of surrounding vehicles inside the ego vehicle's sensor range. Next, the context-generator block processes the past trajectories and provides context-aware grid maps and POMs. The action engine shown in the Fig.\ref{schematic_dia} has been trained in two different ways. At first, an imitation model is learned, then a DDQN model is trained. The imitative model has two parts. The first part consists of spatio-temporal information extraction with a Convolutional Neural Network (CNN), and the second part is a decision neural network. We first train an imitative model for maneuver decisions similar to that given in the DST-CAN \cite{Jayabrata2022DST_CAN}. The imitative model is trained with rule-based decisions. The imitative model first uses a CNN to encode the context-aware social grid maps to a fixed-size state encoding. The CNN uses two three-dimensional convolutional layers. This encoded state is then passed through two different fully connected decision head layers. These decision head layers output lateral and longitudinal decisions. The imitative model is trained with binary cross entropy loss as given in Equ.\ref{eq:BCEloss}.
\begin{equation}
    \mathcal{L} = - \frac{1}{N}\sum_{i=1}^{N}y_ilog(p(y_i))+(1-y_i)log(1-p(y_i))
    \label{eq:BCEloss}
\end{equation}
where $y_i$ is the rule based decision and $p(y_i)$ is the predicted decision. $N$ is the total number of samples.
For learning this model, NGSIM US101 and I80 datasets are employed. Each vehicle's lateral and longitudinal actions are extracted from these datasets. Learning imitative models with a highly imbalanced dataset is very difficult. Learning a good driving model with an imbalanced long-tail dataset like NGSIM is very challenging. Table \ref{data_summary} shows how the different driving decisions are distributed according to traffic rules.
\begin{table}[h]
\centering
\begin{tabular}{lrrr}
                    & \multicolumn{1}{l}{}      & \multicolumn{1}{l}{}      & \multicolumn{1}{l}{}      \\ \hline
                    & \multicolumn{1}{c}{US101} & \multicolumn{1}{c}{US101} & \multicolumn{1}{c}{US101} \\
                    & 07:50-08:05  & 08:05-08:20      & 08:20-08:35          \\ \hline
Follow same lane  & 90.45\%  & 90.51\% & 90.78\% \\
Hard left  &  0.26\%  &  0.27\%  &  0.18\%   \\
Soft left   & 4.13\%  & 4.13\%  & 4.24\%  \\
Hard right & 0.12\%  & 0.20\%  &  0.19\%  \\
Soft right & 5.04\%  & 4.89\% & 4.60\%  \\
Accelerate & 13.65\%  & 14.37\%  & 15.44\% \\
Brake & 0.36\%  & 0.75\%  & 0.84\% \\
Cruise & 79.79\%  & 77.18\%  & 75.36\% \\
Decelerate & 6.20\%  & 7.70\%  & 8.35\% \\
Same lane \& cruise & 70.62\%  & 68\%  & 66.52\% \\
Total samples       &  1180598                  &  1403095              &  1515240              \\ \hline
                    & \multicolumn{1}{l}{}      & \multicolumn{1}{l}{}      & \multicolumn{1}{l}{}      \\ \hline
                    & \multicolumn{1}{c}{I80}   & \multicolumn{1}{c}{I80}   & \multicolumn{1}{c}{I80}   \\
                    & 16:00-16:15               & 17:00-17:15           & 17:15-17:30           \\ \hline
Follow same lane  & 90.55\%  & 91.82\%  & 92.34\%  \\
Hard left  & 0.14\%  & 0.37\%  & 0.52\%  \\
Soft left  & 3.20\%  & 2.69\%  & 2.36\%  \\
Hard right  & 0.47\%  & 1.04\%  & 1.26\%  \\
Soft right  & 5.64\%  & 4.08\%  & 3.53\%  \\
Accelerate  & 14.43\%  & 14.65\%  & 14.88\%  \\
Brake  & 0.80\%  & 2.73\%  & 3.82\%  \\
Cruise  & 75.64\%  & 72.36\%  & 70.97\%  \\
Decelerate  & 9.13\%  & 10.26\%  & 10.33\%  \\
Same lane \& cruise  & 66.79\%  & 65.59\%  & 65.09\%  \\
Total samples       &  1262678                  &  1549918              &  1753791              \\ \hline
\end{tabular}
\caption{Details of NGSIM traffic dataset}
\label{data_summary}
\end{table}
At first, all the data are used for the first training epoch. For some input spatio-temporal grid maps, the imitation network decides the same as rule-based decisions. Those spatio-temporal grid maps are removed from training data for imitation learning. Since most of the dataset comprises cruising on the same lane, a unique data pruning method has been employed for training. Hence, only a tiny portion $(20 \%)$ of cruise maneuvers are taken for training the imitative model after the data pruning process. After training the imitative model, we want to learn how the DDQN-based decision algorithm works for different scenarios. For training the DDQN model, two $Q$ networks (primary network $Q_{\theta}$ and target network $Q_{\theta ^ {'}}$) are used during training. The $Q$ network has been trained for different driving datasets sequentially. The input datasets come sequentially from low-density ($ld$) and high-density ($hd$) datasets for training the $Q$ network. For scenarios where rule-based decisions are cruising, only $50 \%$ of those transitions $(s_t, a_t, r_t, s_{t+1})$ are stored in the replay buffer. For other rule-based decisions, all transitions are stored in the replay buffer. The same trained CNN for the imitative model has been used for state encoding for PMP-DRL. DDQN has been used for action selection. After the state encoding by CNN, three fully connected neural network layers are used for the decision head. The DDQN is trained with the Huber loss function. The final fully connected layer gives the Q values for different decisions. The decision with the highest Q value is taken for an action. Algo. \ref{training procedure} describes the training procedure of the imitative and offline data-driven reinforcement learning-based models in detail. 
\begin{algorithm}
\begin{algorithmic}[1]
\State{\textbf{Part 1}: Imitative model training}
\State{\textbf{Input}: Context-Aware Spatio Temporal Grids as tensors with rule-based maneuver decisions}
\State{\textbf{Initialize}: CNN with the final output of the maneuver action classes and a blank list $\mathcal{L}$ for storing tensor indices, initialize cruise count for data pruning $k_{cruise}=0$}
\State{\textbf{Step 1:} Pass all tensors through model for $1^{st}$ epoch}
\For{$i \leftarrow 1$ to (Total number of social tensors)}
\If{predicted action = actual action}
\State{Save the tensor index $i$ to the list $\mathcal{L}$}
\EndIf
\EndFor
\State{\textbf{Step 2:} Remove the tensors in list $\mathcal{L}$ from training dataset and get new training dataset with remaining data}
\State{\textbf{Step 3:} Train imitative model with new training dataset}
\For{$i \leftarrow 1$ to $\mathcal{E}$ number of epochs}
\For{$j \leftarrow 1$ to (batch size $b$)}
\If {rule-based lateral action = cruise and $k_{cruise} \% 5 = 0$}
\State{$k_{cruise} \leftarrow k_{cruise} +1$ and go to line 19}
\Else
\State{Go to line 19}
\EndIf
\State{Predict the maneuver action classes as probabili \phantom a \phantom . \quad \phantom a  -ties and calculate the Binary Cross Entropy (BCE) \phantom a \phantom a \quad loss $\mathcal{L}$ as given in \ref{eq:BCEloss}} \label{Train_BCE}
\EndFor
\State{Back propagate gradients to learn weights}
\EndFor
\State{\textbf{Part 2:} RL training}
\State{\textbf{Initialize}: Load the pre-trained CNN from the imitative model for state encoding and fix CNN weights. Initialize primary network $Q_{\theta}$ and target network $Q_{\theta ^ {'}}$.}
\For{dataset in $[US101_{ld}, I80_{hd}, I80_{ld}, US101_{hd}]$}
\State{Initialize Replay Buffer $\mathcal{B}; \ k_{cruise} = 0$}
\For {$veh \leftarrow 1 $ to (vehicles available in dataset)}
\State{With probability $\epsilon$, choose $a_t$ randomly}
\State{else choose $a_t = argmax_{a^{'}\in \mathcal{A}} \ Q(s_t, a^{'})$}
\State{Take action $a_t$ and get $s_{t+1}, \ r_{t}$ from environment}
\If {rule-based lateral action = cruise}
\State{$k_{cruise} \leftarrow k_{cruise} +1$}
\If{$k_{cruise} \% 2 = 0$}
\State{store $(s_t,\ a_t, \ r_t,\ s_{t+1})$ in $\mathcal{B}$}
\EndIf
\Else 
\State{store $(s_t,\ a_t, \ r_t,\ s_{t+1})$ in $\mathcal{B}$}
\EndIf
\EndFor
\For{each update step}
\State{sample $e_t = (s_t, a_t, r_t,s_{t+1})$ from ${\mathcal{B}}$}
\State{$Q^*(s_t, a_t) \approx r_t + \gamma Q_{\theta}(s_{t+1}, argmax_{a^{'}}Q_{\theta^{'}}(s_{t+1}, a^{'}))$}
\State{Perform gradient step on $(Q^*(s_t, a_t) - Q_{\theta}(s_t,a_t))^2$}
\EndFor
\State{Update target network $Q_{\theta^{'}}$ parameters}
\EndFor
\caption{Imitative and PMP-DRL training pseudo code}
\label{training procedure}
\end{algorithmic}
\end{algorithm}

\section{Performance evaluation of PMP-DRL}
\label{Performance evaluation of PMP-DRL}
This section presents the performance evaluation of the proposed PMP-DRL method based on real-world NGSIM I80 and US101 \cite{colyar2007us101} datasets. The simulation setup is described first, then performance evaluation metrics are described. PMP-DRL's performance is compared with the rule-based method and recent baseline imitative model DST-CAN \cite{Jayabrata2022DST_CAN}. Finally, ablation studies have been conducted to gain insight into the importance of predictive modules with past context encoding and reward design components, and study results are presented to indicate the effectiveness of the prediction module.

\subsection{RL Simulation setup}
A simulation environment had to be developed where surrounding vehicles show human driving behaviors. This kind of environment can capture real-world mixed autonomous-human driving interactions. If surrounding vehicles follow some hand-coded rules, then autonomous ego vehicles cannot understand human behaviors. Hence, an OpenAI Gym environment has been created based on the actual human-driving NGSIM I-80 and US-101 datasets. The datasets consist of low, medium, and highly congested traffic scenarios. The vehicle trajectories in the datasets are recorded at two different locations at different times of the day. The datasets comprise many traffic scenarios in the study regions, including on-ramps and off-ramps. These different scenarios include significant changes in traffic patterns. Since NGSIM is a real-world dataset, it also has the actual maneuver decisions taken by human drivers. Generally, a vehicle changes lanes within 8 to 10 seconds, as reported in \cite{Song2022LaneChangeReview}. Hence, we consider lane-changing action if a vehicle's lane id changes from the previous 4 seconds to the next 4 seconds. A braking decision is assumed if a vehicle's average speed over the next 5 seconds is lower than 0.8 times the current speed. The environment is designed so that the algorithm controls only one vehicle. Other surrounding vehicles follow the trajectories given in the dataset. One episode is defined for every vehicle's trajectory in the dataset. The low (ld) and highly (hd) congested traffic datasets are used for training the algorithm. After training, the algorithm is tested based on the dataset's unseen medium congestion traffic scenarios. 

\subsection{Performance Evaluation Metrics}
The performance of an autonomous vehicle algorithm is evaluated based on its ability to decide on a safe and comfortable action. The evaluation methodology involves examining the anticipated outcomes of the present action within a short-term horizon. Specifically, the projected position of the ego vehicle after 0.1 seconds is computed based on the currently selected action. The algorithm's performance evaluation encompasses considerations of both safety and passenger comfort. The algorithm's performance is evaluated based on the following metrics:
\begin{itemize}
    \item \textit{Average acceleration}: This metric is derived by computing the mean value of the ego vehicle's acceleration. It serves as a measure of the ego vehicle's ability to navigate traffic scenarios with speed and fluidity. A higher metric value indicates a greater capacity for the vehicle to achieve accelerated movement within the traffic environment.
    \item \textit{Percentage of uncomfortable scenarios}: Passenger comfort is an important measure to evaluate the performance of an autonomous vehicle. To evaluate passenger comfort, we can use the number of times a passenger experiences a jerk. Jerk is a measure of how quickly the vehicle's acceleration changes. There will be no jerk if the algorithm decides on an action that changes smoothly (like accelerating to cruising). However, there will be a jerk if the algorithm decides on an action that changes for more than one smooth change (like accelerating to braking). Jerk is calculated by dividing the number of times the algorithm's action changes for more than one smooth change by the total number of steps. The lower the jerk metric, the better the passenger comfort. Minimizing jerk can improve passenger comfort and make autonomous vehicles more enjoyable.
    \item \textit{Percentage near collision scenarios}: This metric leverages the projected position of the ego vehicle, considering a prediction horizon of 0.1 seconds. It identifies instances where any surrounding vehicle encroaches upon the proximity range associated with near-collision events, as specified in the reward design. These occurrences are tallied as near-collision scenarios, and the average frequency of such events is documented. A lower value of this metric is deemed advantageous, indicative of a reduced incidence of near-collision situations.
\end{itemize}
\begin{figure}
    \centering
    \begin{subfigure}[b]{0.48\textwidth}
        \centering
        \includegraphics[scale=0.25]{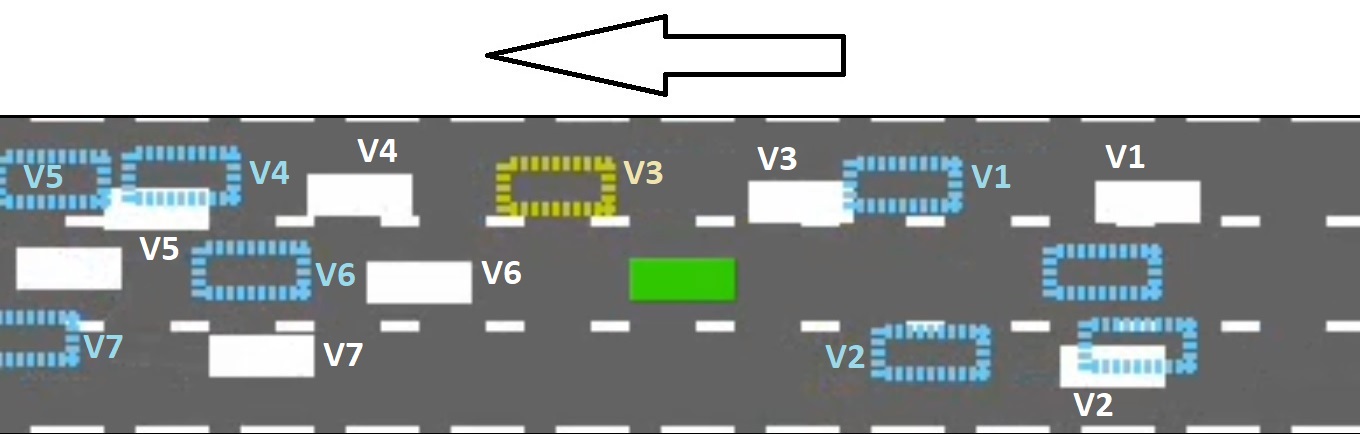}
        \caption{scenario 1}
        \label{scene_1}
    \end{subfigure}
    \begin{subfigure}[b]{0.48\textwidth}
        \centering
        \includegraphics[scale=0.25]{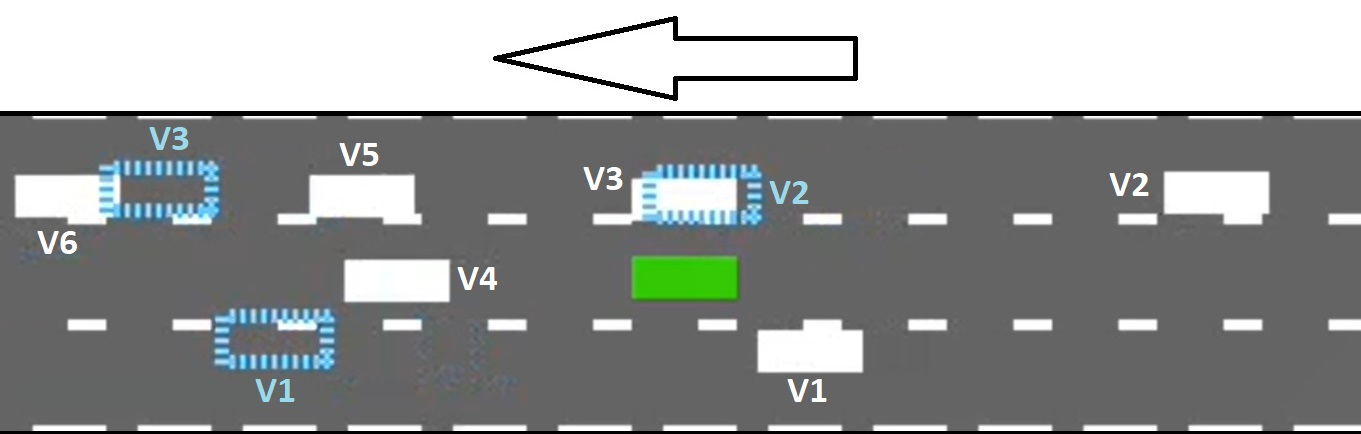}
        \caption{scenario 2}
        \label{scene_2}
    \end{subfigure}
    \caption{Different driving scenarios}
    \label{driving_scenarios}
\end{figure}
Two scenarios are shown in Fig.\ref{driving_scenarios}. The ego vehicle is shown in green. Other surrounding vehicles in the context region of the ego vehicle are shown in white. The near future positions of other surrounding vehicles are given in dotted boxes. The blue dotted boxes are for those surrounding vehicles' future positions that will be at a safe distance from the ego vehicle. Yellow dotted boxes show the vehicles that will be near the ego vehicle. A typical scenario where vehicles $V_2, V_4, V_5, V_6$ are slowing down, is shown in Fig.\ref{scene_1}. Vehicle $V_7$ is changing to the right lane in Fig.\ref{scene_1}. Another scenario where vehicle $V_1$ is changing to the right lane is shown in Fig.\ref{scene_2}.

\subsection{Performance evaluation}
The metrics defined above are used to evaluate the proposed PMP-DRL method's performance. Since some standard traffic rules are already available for driving,  rule-based decisions are also considered for comparison. Also, this study tries to find the effectiveness of the self-learning strategy of data-driven reinforcement learning for autonomous driving. For this purpose, the PMP-DRL results have also been compared with a recent imitative model \cite{Jayabrata2022DST_CAN} based driving. The imitative model was trained using a supervised maneuver classification learning manner. This imitative model also uses the advantages of surrounding vehicles' future position prediction for decision planning. In this study, we have changed the action space of DST-CAN \cite{Jayabrata2022DST_CAN} to match our defined action space. We similarly trained the imitative model with rule-based decisions as ground truth decisions. The training and testing dataset setup has different data distributions of vehicle velocities, as shown in Table \ref{veh_vel_distribution}. 
\begin{table}[h]
\centering
\begin{tabular}{lrrr}
                    & \multicolumn{1}{l}{}
                    & \multicolumn{1}{l}{}      & \multicolumn{1}{l}{}      \\ \hline
                    Vehicle velocity
                    & \multicolumn{1}{c}{US101} & \multicolumn{1}{c}{US101} & \multicolumn{1}{c}{US101} \\
                    ($m/s$)
                    & 07:50-08:05               & 08:05:08:20           & 08:20-08:35          \\ \hline
Mean  & 4.827  & 8.939 & 7.836 \\
Standard deviation  & 3.623  & 4.089 & 3.672   \\
25 percentile   & 9.114  & 6.096  & 5.093  \\
50 percentile & 11.963  & 9.153  & 8.035   \\
75 percentile & 14.438  & 12.168 & 10.638  \\
\hline
                    & \multicolumn{1}{l}{}      & \multicolumn{1}{l}{}      & \multicolumn{1}{l}{}  \\ \hline
                    Vehicle velocity
                    & \multicolumn{1}{c}{I80}   & \multicolumn{1}{c}{I80}   & \multicolumn{1}{c}{I80} \\
                    ($m/s$)
                    & 16:00-16:15 & 17:00-17:15 & 17:15-17:30 \\ \hline
Mean  & 7.722  & 5.622  & 4.827  \\
Standard deviation  & 4.061  & 3.889 & 3.623   \\
25 percentile  & 5.624  & 2.844  & 2.033  \\
50 percentile  & 7.483  & 4.877 & 4.234   \\
75 percentile  & 9.345  & 7.620  & 6.870  \\
\hline
\end{tabular}
\caption{Details of vehicle velocities for NGSIM dataset}
\label{veh_vel_distribution}
\end{table}
Consensus and conflict scenarios are evaluated separately for a comparison with the imitative model. Consensus cases are those where human decisions and rule-based decisions are the same. The PMP-DRL method has been trained in a sequence of datasets described in Algo. \ref{training procedure}. After each training dataset, the performance has been evaluated and shown in Table \ref{training_data_results}. From Table \ref{training_data_results}, it can be found that the driving performance improved for all metrics sequentially. In the beginning, the average acceleration is high ($0.314 m/s^2$ and $0.311 m/s^2$, for consensus and conflict cases, respectively). However, the near-collision scenarios are also high initially ($11.812 \%$ and $19.019 \%$ for consensus and conflict cases, respectively). Also, it can be found that there exists a trade-off between uncomfortable scenarios and near-collision scenarios. From these results, it can be concluded that PMP-DRL can understand driving scenarios better and learn to improve all the metrics.
\begin{table}[]
\centering
\begin{tabular}{|cccc|}
\hline
\multicolumn{1}{|c|}{} &
  \multicolumn{1}{c|}{\begin{tabular}[c]{@{}c@{}}Average\\ acceleration ($m/s^2$)\end{tabular}} &
  \multicolumn{1}{c|}{\begin{tabular}[c]{@{}c@{}}Uncomfortable\\ scenarios (\%)\end{tabular}} &
  \begin{tabular}[c]{@{}c@{}}Near collision\\ (\%)\end{tabular} \\ \hline
\multicolumn{4}{|c|}{Consensus cases}                                                                                                                   \\ \hline
\multicolumn{1}{|c|}{\begin{tabular}[c]{@{}c@{}}US101\\ (7:50-08:05)\end{tabular}}  & \multicolumn{1}{c|}{0.314} & \multicolumn{1}{c|}{37.349} & 11.812 \\ \hline
\multicolumn{1}{|c|}{\begin{tabular}[c]{@{}c@{}}I80\\ (17:15-17:30)\end{tabular}}   & \multicolumn{1}{c|}{0.229} & \multicolumn{1}{c|}{37.532} & 7.538  \\ \hline
\multicolumn{1}{|c|}{\begin{tabular}[c]{@{}c@{}}I80\\ (16:00-16:15)\end{tabular}}   & \multicolumn{1}{c|}{0.267} & \multicolumn{1}{c|}{37.508} & 8.697  \\ \hline
\multicolumn{1}{|c|}{\begin{tabular}[c]{@{}c@{}}US101\\ (08:20-08:35)\end{tabular}} & \multicolumn{1}{c|}{0.269} & \multicolumn{1}{c|}{37.474} & 7.991  \\ \hline
\multicolumn{4}{|c|}{Conflict cases}                                                                                                                    \\ \hline
\multicolumn{1}{|c|}{\begin{tabular}[c]{@{}c@{}}US101\\ (7:50-08:05)\end{tabular}}  & \multicolumn{1}{c|}{0.311} & \multicolumn{1}{c|}{37.675} & 19.019 \\ \hline
\multicolumn{1}{|c|}{\begin{tabular}[c]{@{}c@{}}I80\\ (17:15-17:30)\end{tabular}}   & \multicolumn{1}{c|}{0.242} & \multicolumn{1}{c|}{37.514} & 16.445 \\ \hline
\multicolumn{1}{|c|}{\begin{tabular}[c]{@{}c@{}}I80\\ (16:00-16:15)\end{tabular}}   & \multicolumn{1}{c|}{0.267} & \multicolumn{1}{c|}{37.508} & 8.697  \\ \hline
\multicolumn{1}{|c|}{\begin{tabular}[c]{@{}c@{}}US101\\ (08:20-08:35)\end{tabular}} & \multicolumn{1}{c|}{0.271} & \multicolumn{1}{c|}{37.354} & 13.931 \\ \hline
\end{tabular}
\caption{PMP-DRL performance during training sequence}
\label{training_data_results}
\end{table}
Before presenting the comparative results, we found how many near-collision scenarios are inside the NGSIM dataset according to the near-collision region around the ego vehicle described in the reward structure. The details are shown in Table \ref{human_decision_near_collision}.
\begin{table}[]
\centering
\begin{tabular}{|c|cc|cc|}
\hline
 &
  \multicolumn{2}{c|}{Consensus cases (\%)} &
  \multicolumn{2}{c|}{Conflict cases (\%)} \\ \hline
 &
  \multicolumn{1}{c|}{\begin{tabular}[c]{@{}c@{}}I80\\ 17:00-17:15
  \end{tabular}} &
  \begin{tabular}[c]{@{}c@{}}US101\\ 08:05-08:20
  \end{tabular} &
  \multicolumn{1}{c|}{\begin{tabular}[c]{@{}c@{}}I80\\ 17:00-17:15
  \end{tabular}} &
  \begin{tabular}[c]{@{}c@{}}US101\\ 08:05-08:20
  \end{tabular} \\ \hline
\begin{tabular}[c]{@{}c@{}}Near\\ collision\\ scenarios\end{tabular} &
  \multicolumn{1}{c|}{7.312 \%} &
  2.099 \% &
  \multicolumn{1}{c|}{16.094 \%} &
  4.650 \% \\ \hline
\end{tabular}
\caption{Near collision scenarios inside test dataset}
\label{human_decision_near_collision}
\end{table}
Table \ref{US-101 comparison_table} shows the results of comparing different methods. This table has three results for traffic rule-based (Rule based), PMP with imitation learning (Imitation), and PMP with Deep Reinforcement Learning (PMP-DRL). 
\begin{table}[]
\begin{tabular}{|cccc|}
\hline
\multicolumn{1}{|c|}{} &
  \multicolumn{1}{c|}{\begin{tabular}[c]{@{}c@{}}Average\\ acceleration ($m/s^2$)\end{tabular}} &
  \multicolumn{1}{c|}{\begin{tabular}[c]{@{}c@{}}Uncomfortable \\ scenarios (\%)\end{tabular}} &
  \begin{tabular}[c]{@{}c@{}}Near collision\\ (\%)\end{tabular} \\ \hline
\multicolumn{4}{|c|}{I-80 medium congestion traffic}                                                                          \\ \hline
\multicolumn{4}{|c|}{Consensus cases}                                                                                         \\ \hline
\multicolumn{1}{|c|}{Rule based} & \multicolumn{1}{c|}{0.089}          & \multicolumn{1}{c|}{71.058}  & 7.235           \\ \hline
\multicolumn{1}{|c|}{Imitation}  & \multicolumn{1}{c|}{0.132}          & \multicolumn{1}{c|}{57.325}          & \textbf{7.087}          \\ \hline
\multicolumn{1}{|c|}{PMP-DRL}    & \multicolumn{1}{c|}{\textbf{0.239}} & \multicolumn{1}{c|}{\textbf{37.481}} & {7.142} \\ \hline
\multicolumn{4}{|c|}{Conflict cases}                                                                                          \\ \hline
\multicolumn{1}{|c|}{Rule based} & \multicolumn{1}{c|}{0.070}          & \multicolumn{1}{c|}{77.600}          & {15.435} \\ \hline
\multicolumn{1}{|c|}{Imitation}  & \multicolumn{1}{c|}{0.082}          & \multicolumn{1}{c|}{73.766}          & \textbf{15.396}          \\ \hline
\multicolumn{1}{|c|}{PMP-DRL}    & \multicolumn{1}{c|}{\textbf{0.254}} & \multicolumn{1}{c|}{\textbf{37.330}} & 15.869          \\ \hline
\multicolumn{4}{|c|}{US-101 medium congestion traffic}                                                                        \\ \hline
\multicolumn{4}{|c|}{Consensus cases}                                                                                         \\ \hline
\multicolumn{1}{|c|}{Rule based} & \multicolumn{1}{c|}{0.082}          & \multicolumn{1}{c|}{74.910}          & 9.323          \\ \hline
\multicolumn{1}{|c|}{Imitation}  & \multicolumn{1}{c|}{0.087}          & \multicolumn{1}{c|}{73.245}          & 9.271          \\ \hline
\multicolumn{1}{|c|}{PMP-DRL}    & \multicolumn{1}{c|}{\textbf{0.282}} & \multicolumn{1}{c|}{\textbf{37.553}} & \textbf{9.081} \\ \hline
\multicolumn{4}{|c|}{Conflict cases}                                                                                          \\ \hline
\multicolumn{1}{|c|}{Rule based} & \multicolumn{1}{c|}{0.034}          & \multicolumn{1}{c|}{89.268}          & 14.998          \\ \hline
\multicolumn{1}{|c|}{Imitation}  & \multicolumn{1}{c|}{0.036}           & \multicolumn{1}{c|}{88.678}          & 14.982         \\ \hline
\multicolumn{1}{|c|}{PMP-DRL}    & \multicolumn{1}{c|}{\textbf{0.283}} & \multicolumn{1}{c|}{\textbf{37.474}} & \textbf{14.728} \\ \hline
\end{tabular}
\caption{Comparative results}
\label{US-101 comparison_table}
\end{table}
It shows that the proposed PMP-DRL method performs much better than the other imitative and rule-based models. The imitative model suffers from unseen data distribution. The PMP-DRL-based method decides more safe and more comfortable actions. As the average acceleration metric shows, it also moves faster through the traffic. A trade-off exists between near-collision scenarios and uncomfortable scenarios with average acceleration. For the comfort of passengers, ego vehicle can not change its decision very frequently. If some scenario changes very quickly, the ego vehicle must change its decisions abruptly. This abrupt change will make the ego vehicle safe; however, the comfortability of passengers will be reduced. Table \ref{US-101 comparison_table} shows that the PMP-DRL method performs similarly to the imitative model for near-collision scenarios. It has been found that the PMP-DRL method takes such decisions where $7.142 \%$ and $15.869 \%$ of cases occur in near-collision scenarios for the I80 dataset. In the human-driven decisions, there are $7.132 \%$ and $16.094 \%$ cases; at least one vehicle was in the near collision region as given in Table \ref{human_decision_near_collision}. From this, we can infer that the DRL-PMP method makes decisions such that fewer near-collision scenarios happen during driving than human driving. The imitative model mimics actions that are only safe but uncomfortable. Hence, the imitative model performs poorly compared to PMP-DRL for comfortable scenarios. For comfortability, PMP-DRL gives $19.84\%$ and $35.69\%$ improvement for consensus cases and $36.69\%$ and $51.20\%$ improvement for conflict cases in comparison with the imitative model. Also, the average acceleration is much higher with PMP-DRL than with the imitative model. This implies that the ego vehicle can move smoothly without making too many jerks. 

\subsection{Ablation study}
This section presents the importance of surrounding vehicles' future position prediction module in maneuver planning for AV. The results are presented with the same metrics defined earlier. The input context-aware social grid maps include the past, present, and predicted future context information around the ego vehicle. The different time steps in the contextual grid maps present information that can help take safe, efficient, and comfortable actions. Hence, in this study, three different models have been trained. One model(Imitation(P)) has been trained with only the current instant occupancy grid map. Another model (Imitation(PP)) uses all the 3 seconds of past and current occupancy maps for training. Finally, the proposed model (PMP-DRL) has been trained with predicted future POMs with past occupancy maps. The comparisons are shown in Table \ref{time cimparison}.
\begin{table}[]
\centering
\begin{tabular}{|cccc|}
\hline
\multicolumn{1}{|c|}{} &
  \multicolumn{1}{c|}{\begin{tabular}[c]{@{}c@{}}Average\\ acceleration ($m/s^2$)\end{tabular}} &
  \multicolumn{1}{c|}{\begin{tabular}[c]{@{}c@{}}Uncomfortable \\ scenarios (\%)\end{tabular}} &
  \begin{tabular}[c]{@{}c@{}}Near collision\\ (\%)\end{tabular} \\ \hline
\multicolumn{4}{|c|}{I-80 medium congestion traffic}                                                                                \\ \hline
\multicolumn{4}{|c|}{Consensus cases}                                                                                               \\ \hline
\multicolumn{1}{|c|}{Imitation (P)}  & \multicolumn{1}{c|}{0.005}          & \multicolumn{1}{c|}{95.688}          & 7.252           \\ \hline
\multicolumn{1}{|c|}{Imitation (PP)} & \multicolumn{1}{c|}{0.086}          & \multicolumn{1}{c|}{71.847}          & 7.230           \\ \hline
\multicolumn{1}{|c|}{PMP-DRL}        & \multicolumn{1}{c|}{\textbf{0.239}} & \multicolumn{1}{c|}{\textbf{37.481}} & \textbf{7.142}  \\ \hline
\multicolumn{4}{|c|}{Conflict cases}                                                                                                \\ \hline
\multicolumn{1}{|c|}{Imitation (P)}  & \multicolumn{1}{c|}{0.042}          & \multicolumn{1}{c|}{86.281}          & 15.460          \\ \hline
\multicolumn{1}{|c|}{Imitation (PP)} & \multicolumn{1}{c|}{0.070}          & \multicolumn{1}{c|}{77.561}          & \textbf{15.440} \\ \hline
\multicolumn{1}{|c|}{PMP-DRL}        & \multicolumn{1}{c|}{\textbf{0.254}} & \multicolumn{1}{c|}{\textbf{37.330}} & 15.869          \\ \hline
\multicolumn{4}{|c|}{US-101 medium congestion traffic}                                                                              \\ \hline
\multicolumn{4}{|c|}{Consensus cases}                                                                                               \\ \hline
\multicolumn{1}{|c|}{Imitation (P)}  & \multicolumn{1}{c|}{0.002}          & \multicolumn{1}{c|}{98.645}          & 9.308           \\ \hline
\multicolumn{1}{|c|}{Imitation (PP)} & \multicolumn{1}{c|}{0.079}          & \multicolumn{1}{c|}{75.829}          & 9.322           \\ \hline
\multicolumn{1}{|c|}{PMP-DRL}        & \multicolumn{1}{c|}{\textbf{0.282}} & \multicolumn{1}{c|}{\textbf{37.553}} & \textbf{9.081}  \\ \hline
\multicolumn{4}{|c|}{Conflict cases}                                                                                                \\ \hline
\multicolumn{1}{|c|}{Imitation (P)}  & \multicolumn{1}{c|}{0.013}          & \multicolumn{1}{c|}{95.930}          & 14.978          \\ \hline
\multicolumn{1}{|c|}{Imitation (PP)} & \multicolumn{1}{c|}{0.034}          & \multicolumn{1}{c|}{89.275}          & 14.900          \\ \hline
\multicolumn{1}{|c|}{PMP-DRL}        & \multicolumn{1}{c|}{\textbf{0.283}} & \multicolumn{1}{c|}{\textbf{37.474}} & \textbf{14.728} \\ \hline
\end{tabular}
\caption{Comparisons of the effectiveness of prediction}
\label{time cimparison}
\end{table}
The Table shows that the performance improved significantly using the predicted future information in the context-aware social grid maps. There is $34.37\%$ and $38.276\%$ improvement for consensus scenarios, $40.23\%$ and $51.80\%$ improvement for conflicting scenarios, for comfortable decision making. There is also a significant improvement in the average acceleration and near-collision scenarios when using future prediction. 

\section{Conclusion}
\label{conclusions}
This paper presents a data-driven method for Predictive Maneuver Planning with Deep Reinforcement Learning (PMP-DRL) for safe and comfortable autonomous driving. The surrounding vehicles' past trajectories are encoded in an occupancy grid map. Occupancy grid maps are created with the predicted future positions of surrounding vehicles along with their uncertainty estimates. PMP-DRL shows better performance than rule-based and imitative models. The rule-based method looks for safety, not passengers' comfort. The imitative model suffers from uncomfortable decision-making as it tries to learn traffic rules broadly. The results show an improvement with DRL-PMP in near-collision scenarios and a significant improvement of $35.69\%$ and $51.20\%$ in comfortability compared to the imitative model. Also, we show how predictions help in safe and comfortable maneuvering decision-making. There are $40.23\%$ and $51.80\%$ improvements in comfort while using future predictions for maneuver planning. The results clearly indicate the importance of the proposed prediction module in data-driven reinforcement learning over an imitative model.

\bibliographystyle{IEEEtran}
\bibliography{NGSIM-RL-PMP}

\end{document}